\documentclass[conference]{IEEEtran}
\IEEEoverridecommandlockouts
\usepackage{cite}
\usepackage{amsmath,amssymb,amsfonts}
\usepackage{algorithmic}
\usepackage{graphicx}
\usepackage{booktabs}
\usepackage{textcomp}
\usepackage{xcolor}
\usepackage{multirow}

\makeatletter
\newcommand*{\rom}[1]{\expandafter\@slowromancap\romannumeral #1@}
\makeatother

\def\BibTeX{{\rm B\kern-.05em{\sc i\kern-.025em b}\kern-.08em
    T\kern-.1667em\lower.7ex\hbox{E}\kern-.125emX}}
\begin{document}

\title{Predicting the Transition from Short-term to Long-term Memory based on Deep Neural Network\\

\author{\IEEEauthorblockN{Gi-Hwan Shin}
\IEEEauthorblockA{\textit{Dept. Brain and Cognitive Engineering} \\
\textit{Korea University}\\
Seoul, Republic of Korea \\
gh\_shin@korea.ac.kr}
\and
\IEEEauthorblockN{Young-Seok Kweon}
\IEEEauthorblockA{\textit{Dept. Brain and Cognitive Engineering} \\
\textit{Korea University}\\
Seoul, Republic of Korea \\
youngseokkweon@korea.ac.kr}
\and
\IEEEauthorblockN{Minji Lee}
\IEEEauthorblockA{\textit{Dept. Brain and Cognitive Engineering} \\
\textit{Korea University}\\
Seoul, Republic of Korea \\
minjilee@korea.ac.kr}
}

\thanks{20xx IEEE. Personal use of this material is permitted. Permission
from IEEE must be obtained for all other uses, in any current or future media, including reprinting/republishing this material for advertising or promotional purposes, creating new collective works, for resale or redistribution to servers or lists, or reuse of any copyrighted component of this work in other works.}

\thanks{This work was partly supported by Institute for Information \& communications Technology Planning \& Evaluation (IITP) grant funded by the Korea government (MSIT) (No. 2017-0-00451, Development of BCI based Brain and Cognitive Computing Technology for Recognizing User’s Intentions using Deep Learning) and (No. 2019-0-00079, Artificial Intelligence Graduate School Program (Korea University)).}
}

\maketitle

\begin{abstract}
Memory is an essential element in people's daily life based on experience. So far, many studies have analyzed electroencephalogram (EEG) signals at encoding to predict later remembered items, but few studies have predicted long-term memory only with EEG signals of successful short-term memory. Therefore, we aim to predict long-term memory using deep neural networks. In specific, the spectral power of the EEG signals of remembered items in short-term memory was calculated and inputted to the multilayer perceptron (MLP) and convolutional neural network (CNN) classifiers to predict long-term memory. Seventeen participants performed visuo-spatial memory task consisting of picture and location memory in the order of encoding, immediate retrieval (short-term memory), and delayed retrieval (long-term memory). We applied leave-one-subject-out cross-validation to evaluate the predictive models. As a result, the picture memory showed the highest $\kappa$-value of 0.19 ($\pm$0.25) on CNN, and location memory showed the highest $\kappa$-value of 0.32 ($\pm$0.35) in MLP. These results showed that long-term memory can be predicted with measured EEG signals during short-term memory, which improves learning efficiency and helps people with memory and cognitive impairments.
\end{abstract}

\begin{IEEEkeywords}
\textit{long-term memory, short-term memory, electroencephalogram, deep neural network}
\end{IEEEkeywords}

\section{Introduction}
Memory is an important building block in learning and decision-making, and it is a cognitive process that encoding and retrieving new information into the brain \cite{small2001circuit, kalafatovich2020prediction}. Specifically, encoding refers to the initial experience of recognizing and learning information, and retrieval refers to the mental process of searching previously learned information \cite{noh2018single}. In addition, memory is divided into short-term and long-term memory mechanisms according to the passage of time, and these mechanisms have different neurophysiological activity patterns \cite{fukuda2017visual}.

Many studies have used electroencephalogram (EEG) to investigate the people's brain mechanisms associated with memory processes \cite{jo2020prediction}. EEG is widely used in disease diagnosis \cite{amezquita2019novel}, rehabilitation \cite{kim2016commanding, lee2018high}, and brain-computer interface (BCI) \cite{won2017motion, park2016movement, chen2016high} for its practical advantages such as non-invasiveness and relatively inexpensive devices. However, EEG signals are dimensional and complex because it is based on a time series of events sampled with high temporal resolution and distributed spatially across multiple scalp locations \cite{johannesen2016machine, lee2015subject}. Therefore, there is an increased need for computational frameworks that can mine large amounts of data to identify the features of the EEG signals most relevant to tasks.

Recently, studies were conducted to predict short- and long-term memory by extracting features of EEG signals when encoding using deep neural networks such as multilayer perceptron (MLP) and convolutional neural network (CNN). For example, Sun \textit{et al}. \cite{sun2016remembered} proposed a convEEGNN to predict whether an item will be remembered or forgotten after encoding, resulting in an average prediction accuracy of 72.07\%. In addition, Kang \textit{et al}. \cite{kang2020eeg} predicted long-term memory performance using various classification models such as linear discriminant analysis and CNN. These results showed that EEG signals during encoding can predict the performance of subsequent memory effects. However, few studies have predicted the performance of long-term memory using EEG signals of successful short-term memory.

In this study, we aimed to predict the long-term memory using the features of the EEG signals of remembered items in short-term memory. Participants performed an encoding, immediate retrieval (short-term memory), and delayed retrieval (long-term memory) of visuo-spatial memory task consisting of picture and location memory. We used basic deep neural networks to investigate whether long-term memory was predicted. Our results showed the feasibility to improve learning efficiency and help people with memory and cognitive impairments.

\begin{figure*}[t!]
\centering
\scriptsize
\includegraphics[width=\textwidth]{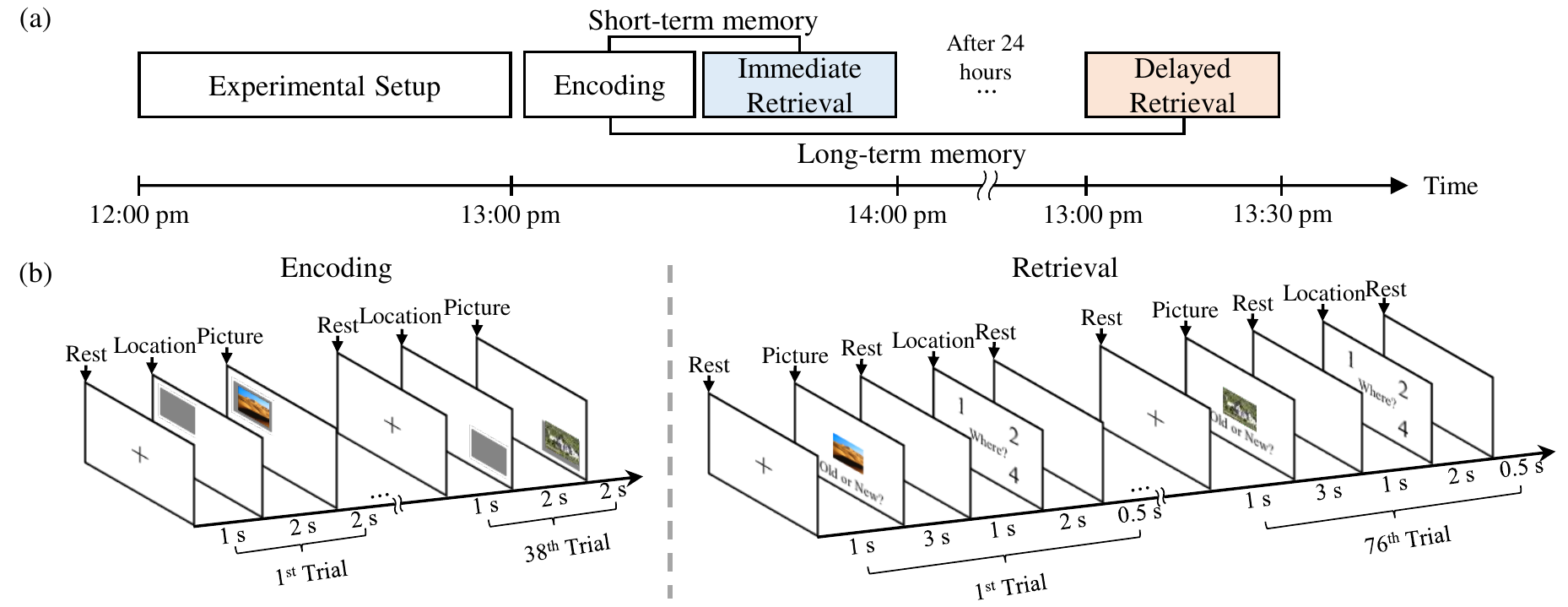}
\caption{Study design. (a) The experimental procedure consisted of encoding, immediate retrieval (short-term memory), and delayed retrieval (long-term memory) after the experimental setup. (b) The visuo-spaital memory tasks consisting of picture memory and location memory were displayed on the screen with the encoding and retrieval of 38 trials and 76 trials, respectively.}
\end{figure*}

\section{Materials and Methods}
\subsection{Participants and Experimental Procedure}
We recorded EEG data from seventeen healthy participants (4 females; 25.8 $\pm$ 1.6 years old) with no history of neurological and psychiatric disorders. This study was approved by the Institutional Review Board at Korea University (KUIRB-2020-0112-01), and each participant gave written informed consent before the experiments.

Participants arrived at the laboratory at 12:00 pm. Following preparation for EEG recording, they performed encoding as initial learning of the memory task. Subsequently, an immediate retrieval for short-term memory was conducted. After 24 hours, a delayed retrieval corresponding to long-term memory was performed (Fig. 1a).

\subsection{Visuo-spatial Memory Task}
The visuo-spatial memory task consisted of picture and location memory \cite{ladenbauer2016brain}) and was implemented using Psychtoolbox (http://psychtoolbox.org). All participants were asked to encode a set of 38 images (objects, plants, and scenes taken from the SUN database \cite{xiao2010sun}) (picture memory) and where the images were presented (location memory).

In encoding, a fixation cross was displayed on the screen for 1 s and a gray square randomly appears at one of the quadrants for 2 s. After that, the picture was presented for 2 s on the gray square. In immediate and delayed retrieval, 38 encoded images and 38 new images were presented in random order. A fixation cross was displayed for 1 s, and each participant pressed a key (with right hand on main keyboard; o - old or n - new) to recognize whether or not they had viewed the image displayed in the center of the screen within 3 s. For the ``o'' button response, they selected by pressing the key corresponding to the quadrant (with left hand on main keyboard; 1, 2, 3, or 4) that they thought the picture was presented (Fig. 1b).

As a measure of successful short-term memory, the correct answer for each participant was determined as follows: picture memory is the number of correct old responses (hits), and location memory is the correct retrieval of hits. The measure of successful long-term memory is to recognize the same remembered items as successful short-term memory.

\begin{figure*}[t!]
\centering
\scriptsize
\includegraphics[width=\textwidth]{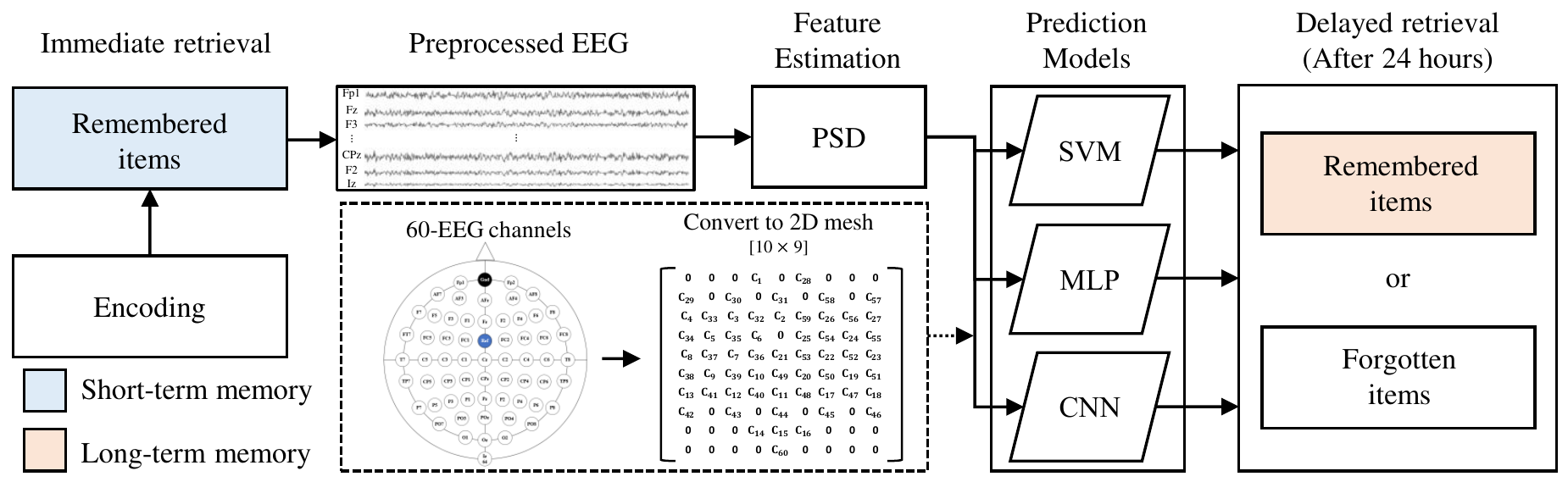}
\caption{An overall framework for predicting long-term memory by using memorized short-term memory. The blue box represents short-term memory, and the orange box represents long-term memory. The square indicated by the gray dotted line means converting the multi-channel EEG signals into a 2-dimensional mesh form when inputting the CNN model. PSD = power spectral density, SVM = support vector machine, MLP = multilayer perceptron, CNN = convolutional neural network.}
\end{figure*}

\subsection{EEG Data Acquisition and Processing}
The EEG was recorded using a 60 channels Ag/AgCl electrode system (ActiCap, Brain Products, Germany) with a 1,000 Hz sampling frequency. The electrodes were positioned according to the international 10-20 electrode system. In addition, the reference and ground electrodes were placed on FCz and AFz, respectively. The impedance of all electrodes was kept below 10 k$\Omega$. 

The raw EEG data were processed with MATLAB R2018b using the EEGLAB toolbox \cite{delorme2004eeglab}. Data were down-sampled to 250 Hz, band-pass filtered between 0.5 to 50 Hz, and re-referenced to the average reference.
To analyze the frequency domain of the prepossessed EEG signals, we calculated the power spectral density (PSD) using a fast Fourier transform (FFT) \cite{bulthoff2003biologically}. We divided into 6 frequency bands as follows: delta (0.5-4 Hz), theta (4-7 Hz), alpha (7-12 Hz), spindle (12-15 Hz), beta (15-30 Hz), and gamma (30-50 Hz) bands. PSD was obtained for each frequency component composing those EEG signals \cite{suk2014predicting, lee2017network, lee2020frontal}:

\begin{equation} PSD_{f_1-f_2} = 10*log_{10}(2\int_{f_1}^{f_2} |\hat{x}(2\pi f)|^2 df) \end{equation}

\noindent
where $f_1$, $f_2$ represent the lower and upper frequencies respectively, and $\hat{x}(2\pi f)$ was obtained by FFT. $10*log_{10}(\bullet)$ denotes unit conversion from microvolts to decibels.

\subsection{Predictive Models}
Fig. 2 showed the overall framework for this study. As the input data of the predictive models, we concatenated the PSD values of the items that were remembered during the short-term memory of all participants. In specific, the shape of 6 frequency bands and 60 channels were reshaped and entered into support vector machine (SVM) and MLP classifiers in 2-dimensions (trials $\times$ 360).
In the baseline classifier, we used radial basis function-kernel SVM \cite{suk2011subject}.
In the MLP approach, it consisted of 3 fully connected layers. We used a rectified linear unit (ReLU) as the activation function and a dropout (\textit{p} = 0.3) was applied to avoid overfitting. Finally, two classes of remembered and forgotten items were classified by softmax activation in a fully connected layer. To train this network, we used a batch size of 20 for 50 epochs and cross-entropy with $10^{-5}$ learning rate.
In the CNN model, multiple EEG channels mapped in a 2D mesh form (10 $\times$ 9) to maximize spatial information of brain activity \cite{zhang2018cascade}. The mesh point which is not allocated for the EEG channels is assigned to zeros. Therefore, the inputs of the CNN was organized in 4-dimensions (trials $\times$ PSD $\times$ 2D mesh). The CNN architecture consisted of 2 convolution layers, 1 max-pooling layer, and 1 fully connected layer (Table \rom{1}). We applied batch normalization after each convolution layer to normalize the variance of the learned data. Other than that, it proceeds in the same structure as MLP.

\begin{table}[t!]
\caption{Description of CNN Architecture}
\resizebox{\columnwidth}{!}{
\scriptsize
\renewcommand{\arraystretch}{1.2}
\begin{tabular}{@{\extracolsep{\fill}\quad}cccc}
\hline
\textbf{Layer}     & \textbf{Operation}                                                                   & \textbf{Size   of Feature map} & \textbf{Kernel size}     \\ \hline
0                  & Input                                                                                & 6 $\times$    10 $\times$  9                   & -                        \\
\multirow{2}{*}{1} & \multirow{2}{*}{\begin{tabular}[c]{@{}c@{}}Convolution\\ Batch Normalization\end{tabular}} & \multirow{2}{*}{8 $\times$    8 $\times$  7}   & \multirow{2}{*}{3 $\times$    3} \\
                   &                                                                                      &                                &                          \\
\multirow{2}{*}{2} & \multirow{2}{*}{\begin{tabular}[c]{@{}c@{}}Convolution\\ Normalization\end{tabular}}         & \multirow{2}{*}{64   $\times$  6 $\times$  5}  & \multirow{2}{*}{3 $\times$    3} \\
                   &                                                                                      &                                &                          \\
\multirow{2}{*}{3} & \multirow{2}{*}{\begin{tabular}[c]{@{}c@{}}Max-pooling\\ Dropout (0.3)
\end{tabular}}       & \multirow{2}{*}{64 $\times$  3 $\times$  2}    & \multirow{2}{*}{1 $\times$    2} \\
                   &                                                                                      &                                &                          \\
4                  & Flatten                                                                              & 384                            & -                        \\
5                  & Fully-Connected                                                                      & 2                              & -                \\ \hline           
\end{tabular}
}
\end{table}

\begin{figure}[t!]
\centering
\scriptsize
\includegraphics[width=\columnwidth]{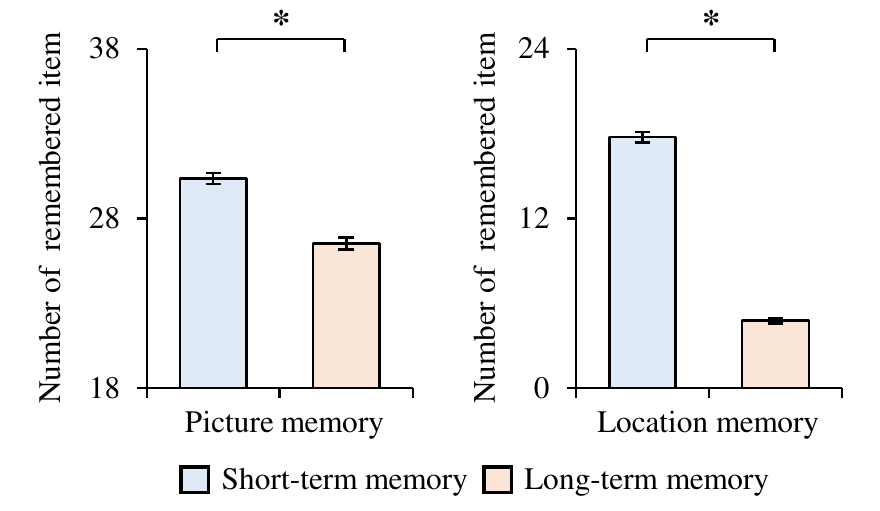}
\caption{The number of items remembered during the retrieval of short- and long-term memory for each memory task. Error bars show standard errors. * indicates statistical significance at \textit{p} $<$ 0.05 determined by \textit{t}-test.}
\end{figure}

\begin{figure*}[t!]
\centering
\scriptsize
\includegraphics[width=\textwidth]{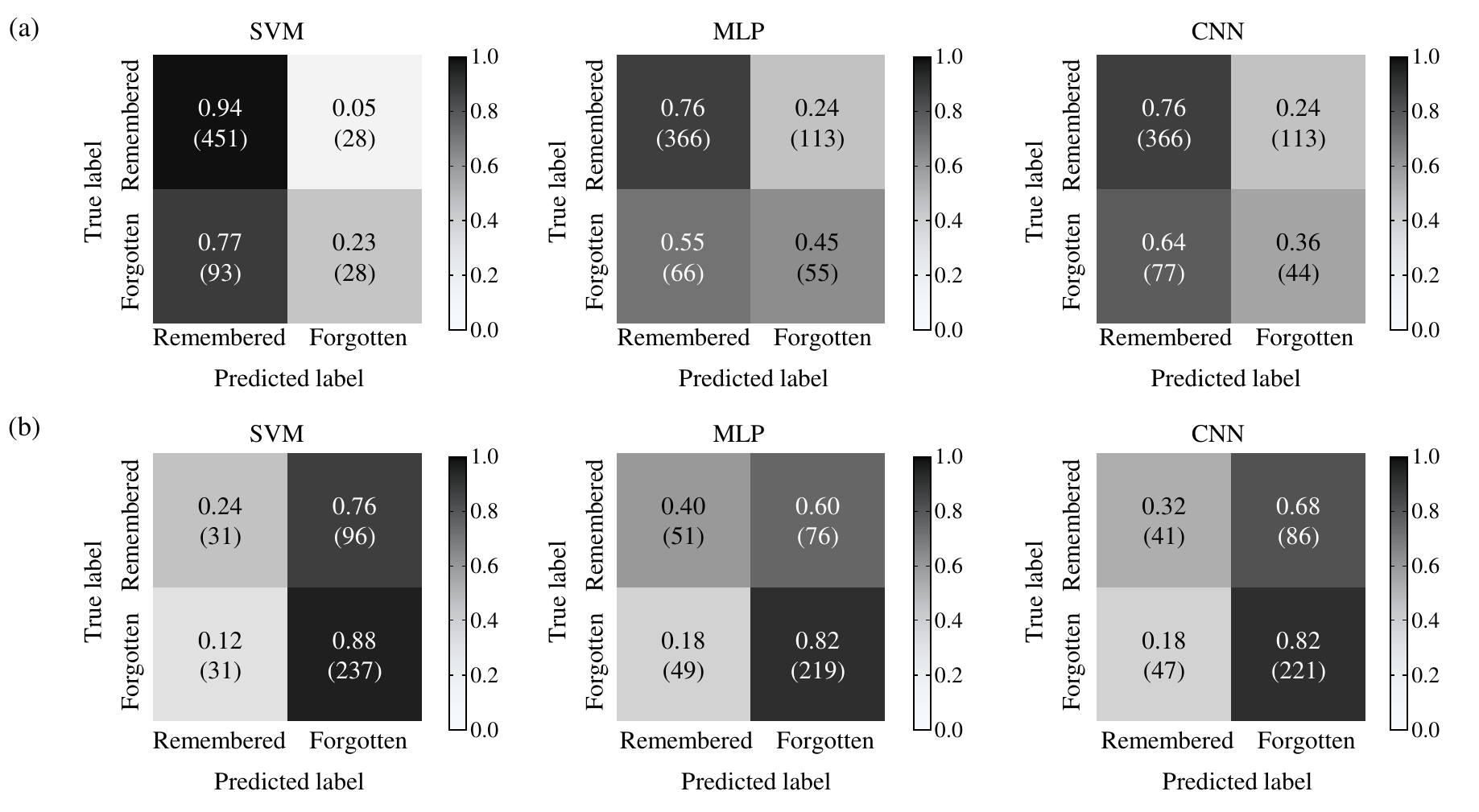}
\caption{The normalized confusion matrix of (a) picture memory and (b) location memory using prediction models (SVM, MLP, and CNN). The row represents the true label and the column represents the predicted label by the predictive model. SVM = support vector machine, MLP = multilayer perceptron, CNN = convolutional neural network.}
\end{figure*}

\subsection{Performance Metrics}
For evaluation of the predictive models, we applied leave-one-subject-out (LOSO) cross-validation. It means taking the data from one participant as the testing set and data from the other remaining participants as the training set. The model's performance was calculated by averaging the values of each cross-validation fold.

We quantified the performance of the classifier models using Cohen's kappa, calculated between the truth label and predicted label. Cohen kappa rescales observed correspondence with that expected from chance and outputting values upper bounded by 1 \cite{cohen1960coefficient}:  

\begin{equation} \kappa\equiv\displaystyle \frac{p_o-p_e}{1-p_e} \end{equation}

\noindent
where ${P_0}$ denotes the accuracy of predictions and ${P_e}$ denotes the probability of coincidence between truth label and predicted label.

We also compared the performance of the models with a 2 $\times$ 2 confusion matrix that expressed four measures: true remembered item, true forgotten item, false remembered item, and false forgotten item. 

\section{Results}
\subsection{Successful Visuo-spatial Memory} 
Fig. 3 showed the average and standard error number of remembered items for all participants in each memory task. In the two memories, short-term memory was 30.35 ($\pm$ 0.31) and 17.76 ($\pm$ 0.37), respectively, and long-term memory was 26.53 ($\pm$ 0.35) and 3.82 ($\pm$ 0.17), respectively. In proportions of summing the memory items, the ratio of remembered and forgotten items in picture memory was 87\% to 13\%, and the ratio of remembered and forgotten items in location memory was 21\% to 79\%.

We also used a paired-\textit{t}test to perform statistical comparisons between short-term memory and long-term memory. As a result, both memories were significantly decreased (picture memory: \textit{t} = -6.86, \textit{p} $<$ 0.001; location memory: \textit{t} = -12.17, \textit{p} $<$ 0.001).

\subsection{Prediction Model of Long-term Memory}
We investigated the predictability of long-term memory using three predictive models. Table \rom{2} indicated the accuracy and $\kappa$-values of the predictive models for the two memory tasks. In picture memory, the results $\kappa$-value of SVM, MLP, and CNN were showed 0.06 ($\pm$ 0.24), 0.15 ($\pm$ 0.25), and 0.19 ($\pm$ 0.25), respectively. Both the accuracy and $\kappa$-values showed higher on CNN than other models. In the location memory, $\kappa$-value of SVM, MLP, and CNN were showed 0.18 ($\pm$ 0.38), 0.32 ($\pm$ 0.35), and 0.24 ($\pm$ 0.32), respectively. Of these, it was the highest in MLP. Also, we confirmed that the standard deviation of accuracy and $\kappa$-value was quite large. This is because the data is unbalanced in both memories, so the data is biased towards the majority class.

Fig. 4 presented a normalized confusion matrix for the two memory tasks. The confusion matrices were computed as the sum of the epochs with the largest $\kappa$-values from a LOSO cross-validation for each subject. The numbers in the diagonal lines represent the percentage of correct classification, and the other numbers represent the percentage of misclassification. As a result, in picture memory, the remembered items of the true label, and the remembered items by the prediction model was well classified. On the other hand, forgotten items of the true label, and the remembered items by the prediction model was misclassified. In the location memory, the forgotten items of the true label and forgotten model of the predicted label were well classified, whereas the remembered items of the true label and the forgotten items of the predicted label were misclassified.

\begin{table}[t!]
\caption{Performance of Predictive Models for Long-term Memory in Each Memory Task}
\resizebox{\columnwidth}{!}{
\scriptsize
\renewcommand{\arraystretch}{1.2}
\begin{tabular}{@{\extracolsep{\fill}\quad}lcccc}
\hline
\textbf{Method} & \multicolumn{2}{c}{\textbf{Picture memory}}                                                                     & \multicolumn{2}{c}{\textbf{Location memory}}                                                                     \\ \cline{2-5}
                & \textbf{Accuracy}                                       & \textbf{$\kappa$-value}                                        & \textbf{Accuracy}                                       & \textbf{$\kappa$-value}                                         \\ \hline
SVM             & \begin{tabular}[c]{@{}c@{}}0.87 \\ ($\pm$0.09)\end{tabular} & \begin{tabular}[c]{@{}c@{}}0.06\\ ($\pm$0.24)\end{tabular} & \begin{tabular}[c]{@{}c@{}}0.80\\ ($\pm$0.01)\end{tabular} & \begin{tabular}[c]{@{}c@{}}0.18\\ ($\pm$0.38)\end{tabular} \\ 
MLP             & \begin{tabular}[c]{@{}c@{}}0.87\\ ($\pm$0.09)\end{tabular}  & \begin{tabular}[c]{@{}c@{}}0.15\\ ($\pm$0.25)\end{tabular} & \begin{tabular}[c]{@{}c@{}}0.82\\ ($\pm$0.01)\end{tabular}  & \begin{tabular}[c]{@{}c@{}}0.32\\ ($\pm$0.35)\end{tabular}  \\
CNN             & \begin{tabular}[c]{@{}c@{}}0.88\\ ($\pm$0.08)\end{tabular}  & \begin{tabular}[c]{@{}c@{}}0.19\\ ($\pm$0.25)\end{tabular} & \begin{tabular}[c]{@{}c@{}}0.74\\ ($\pm$0.02)\end{tabular}  & \begin{tabular}[c]{@{}c@{}}0.24\\ ($\pm$0.32)\end{tabular}             \\ \hline
\end{tabular}
\label{tab1}
}
\end{table}

\section{Discussion}
In this study, EEG signals of remembered items of short-term memory was used as an input to a deep neural network to predict long-term memory. This used the EEG recorded during an immediate retrieval of visuo-spatial memory task. In an experiment using LOSO cross-validation, picture memory showed the highest $\kappa$-value on CNN, and location memory showed the highest $\kappa$-value on MLP. 

Our results suggested the possibility to predict long-term memory with the feature of the  EEG signals of short-term memory using basic deep neural networks. This showed that there is a difference in EEG signals between remembered and forgotten items, similar to existing literature that studied subsequent memory effects \cite{sun2016remembered, chakravarty2020predicting}. Cowan \cite{cowan2008differences} noted that the mechanism of short-term memory is separate but closely related to the mechanism of long-term memory. In memory tasks, the participant was likely to recycle remembered items and triggers similar perceptual and semantic processes in the brain \cite{cameron2005long}. Therefore, we can confirm that long-term memory is activated and predictable by short-term memory.

There are some limitations to thesis results. First, the number of trials to the memory task as input to the predictive model was small. Second, there is a data imbalance problem in classification. In future work, we will increase the number of samples through further experiments and perform re-sampling to solve the problem of data imbalance \cite{roy2019deep}. 

\section{Conclusion}
We predicted long-term memory using only the features of the EEG signals of successful short-term memory. Based on these results, it can be applied to various applications related to memory. For example, it can improve the efficiency of learning effects in terms of education. It may also help diagnose and treat diseases related to memory symptoms such as Alzheimer's and cognitive impairment.

\bibliographystyle{IEEEtran}
\bibliography{REFERENCE}
\end{document}